\documentclass[runningheads]{llncs}
\usepackage{graphicx}
\usepackage{comment}
\usepackage{amsmath,amssymb} 
\usepackage{color}
\usepackage{multirow}
\usepackage{subfigure}
\usepackage{booktabs}
\usepackage{pgfplots}

\usepackage[hidelinks]{hyperref}

\usepackage{xcolor}

\begin{document}
\newcommand{\Rset}{\mathbb{R}}
\newcommand{\Esp}{\mathbb{E}}
\newcommand{\Var}{\mathbb{V}}

\newcommand{\bfM}{\mathbf{M}}
\newcommand{\bfA}{\mathbf{A}}
\newcommand{\bfP}{\mathbf{P}}
\newcommand{\bfF}{\mathbf{F}}
\newcommand{\bfR}{\mathbf{R}}
\newcommand{\bfT}{\mathbf{T}}
\newcommand{\bfS}{\mathbf{S}}
\newcommand{\bfK}{\mathbf{K}}
\newcommand{\bfD}{\mathbf{D}}
\newcommand{\bfI}{\mathbf{I}}
\newcommand{\bfQ}{\mathbf{Q}}

\newcommand{\avec}{\boldsymbol{a}}
\newcommand{\bvec}{\boldsymbol{b}}
\newcommand{\dvec}{\boldsymbol{d}}
\newcommand{\evec}{\boldsymbol{e}}
\newcommand{\hvec}{\boldsymbol{h}}
\newcommand{\xvec}{\boldsymbol{x}}
\newcommand{\yvec}{\boldsymbol{y}}
\newcommand{\uvec}{\boldsymbol{u}}
\newcommand{\vvec}{\boldsymbol{v}}
\newcommand{\pvec}{\boldsymbol{p}}
\newcommand{\gvec}{\boldsymbol{g}}
\newcommand{\svec}{\boldsymbol{s}}
\newcommand{\zvec}{\boldsymbol{z}}

\newcommand{\weights}{\boldsymbol{w}}

\newcommand{\net}{N}
\newcommand{\visionnet}{\net_v}
\newcommand{\audionet}{\net_a}
\newcommand{\visionnetinit}{\visionnet^{(0)}}
\newcommand{\audionetinit}{\audionet^{(0)}}
\newcommand{\cls}{f}

\newcommand{\norm}[2][2]{\left\|#2\right\|_{#1}}

\pagestyle{headings}
\mainmatter

\title{Cross-Task Transfer for Geotagged Audiovisual Aerial Scene Recognition} 

\titlerunning{Cross-Task Transfer for Geotagged Audiovisual Aerial Scene Recognition}
%
\author{Di Hu \inst{1} \and
Xuhong Li\inst{1} \and
Lichao Mou\inst{2,3} \and
Pu Jin\inst{2} \and
Dong Chen\inst{4} \and
Liping Jing\inst{4} \and
Xiaoxiang Zhu\inst{2,3} \and
Dejing Dou\inst{1}\thanks{Corresponding Author}}
\authorrunning{D. Hu et al.}
%
\institute{
Big Data Laboratory, Baidu Research\\
\email{\{hudi04,lixuhong,doudejing\}@baidu.com} \and
Technical University of Munich\\
\email{\{lichao.mou,pu.jin\}@tum.de} \and
German Aerospace Center \\
\email{\{lichao.mou,xiaoxiang.zhu\}@dlr.de} \and
Beijing Key Lab of Traffic Data Analysis and Mining, Beijing Jiaotong University \\
\email{\{chendong,lpjing\}@bjtu.edu.cn}
}

\maketitle

\begin{abstract}
Aerial scene recognition is a fundamental task in remote sensing and has recently received increased interest.
While the visual information from overhead images with powerful models and efficient algorithms yields considerable performance on scene recognition, it still suffers from the variation of ground objects, lighting conditions etc. Inspired by the multi-channel perception theory in cognition science, in this paper, for improving the performance on the aerial scene recognition, we explore a novel audiovisual aerial scene recognition task using both images and sounds as input.
Based on an observation that some specific sound events are more likely to be heard at a given geographic location, we propose to exploit the knowledge from the sound events to improve the performance on the aerial scene recognition.
For this purpose, we have constructed a new dataset named \textit{AuDio Visual Aerial sceNe reCognition datasEt} (ADVANCE).
With the help of this dataset, we evaluate three proposed approaches for transferring the sound event knowledge to the aerial scene recognition task in a multimodal learning framework, and show the benefit of exploiting the audio information for the aerial scene recognition. 
The source code is publicly available for reproducibility purposes.\footnote{\url{https://github.com/DTaoo/Multimodal-Aerial-Scene-Recognition}}


\end{abstract}

\section{Introduction}
Scene recognition is a longstanding, hallmark problem in the field of computer vision, and it refers to assigning a scene-level label to an image based on its overall contents. 
Most scene recognition approaches in the community make use of ground images and have achieved remarkable performance. By contrast, overhead images usually cover larger geographical areas and are capable of offering more comprehensive information from a bird’s eye view than ground images. Hence aerial scene recognition has received increased interest. The success of current state-of-the-art aerial scene understanding models can be attributed to the development of novel convolutional neural networks (CNNs) that aim at learning good visual representations from images.

\begin{figure*}[t]
	\centering
 	\includegraphics[width = \linewidth]{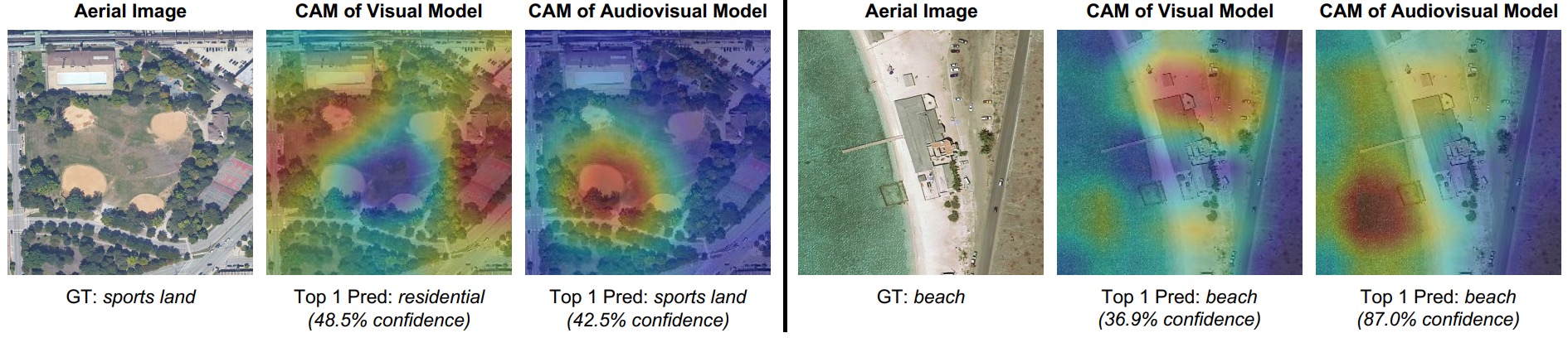}
	\caption{Two examples showing aerial scenes’ soundscapes could be a helpful cue for identifying their scene categories. More details of the audiovisual model please refer to Section~\ref{subsection:bayes}. Here, we use the class activation mapping (CAM) technique to visualize what models are looking.}
	\label{figure_examples}
\end{figure*}

Albeit successful, these models may not work well in some cases, particularly when they are directly used in worldwide applications, suffering the pervasive factors, such as different remote imaging sensors, lighting conditions, orientations, and seasonal variations. A study in neurobiology reveals that human perception usually benefits from the integration of both visual and auditory knowledge. Inspired by this investigation, we argue that aerial scenes’ soundscapes are partially free of the aforementioned factors and can be a helpful cue for identifying scene categories (Fig.~\ref{figure_examples}). This is based on an observation that the visual appearance of an aerial scene and its soundscape are closely connected. For instance, sound events like broadcasting, people talking, and perhaps whistling are likely to be heard in all train stations in the world, and cheering and shouting are expected to hear in most sports lands. However, incorporating the sound knowledge into a visual aerial scene recognition model and assessing its contributions to this task still remain underexplored. In addition, it is worth mentioning that with the now widespread availability of smartphones, wearable devices, and audio sharing platforms, geotagged audio data have been easily accessible, which enables us to explore the topic in this paper.

In this work, we are interested in the audiovisual aerial scene recognition task that simultaneously uses both visual and audio messages to identify the scene of a geographical region. 
To this end, we construct a new dataset, named \textit{AuDio Visual Aerial sceNe reCognition datasEt} (ADVANCE), providing 5075 paired images and sound clips categorized to 13 scenes, which will be introduced in Section~\ref{section:dataset}, for exploring the aerial scene recognition task.
According to our preliminary experiments, simply concatenating representations from the two modalities is not helpful, slightly degrading the recognition performance compared to using a vision-based model.
Knowing that sound events are related to scenes, this preliminary result indicates that the model cannot directly learn the underlying relation between the sound events and the scenes.
So directly transferring the sound event knowledge to scene recognition may be the key to making progress.
Following this direction, with the multimodal representations, we propose three approaches that can effectively exploit the audio knowledge to solve the aerial scene recognition task, which will be detailed in Section~\ref{section:method}.
We compare our proposed approaches with baselines in Section \ref{section:exp}, showing the benefit of exploiting the sound event knowledge for the aerial scene recognition task.

Thereby, this work’s contributions are threefold.
\begin{itemize}
\item The audiovisual perception of human beings gives us an incentive to investigate a novel audiovisual aerial scene recognition task. We are not aware of any previous work exploring this topic.
\item We create an annotated dataset consisting of 5075 geotagged aerial image-sound pairs involving 13 scene classes. This dataset covers a large variety of scenes from across the world.
\item We propose three approaches to exploit the audio knowledge, {\em i.e.}, preserving the capacity of recognizing sound events, constructing a mutual representation in order to learn the underlying relation between sound events and scenes, and directly learning this relation through the posterior probabilities of sound events given a scene. In addition, we validate the effectiveness of these approaches through extensive ablation studies and experiments.
\end{itemize}

\section{Related work}


In this section, we briefly review some related works in aerial scene recognition, multimodal learning, and cross-task transfer.

\paragraph{Aerial Scene Recognition.} Earlier studies on aerial scene recognition \cite{yang2008comparing,risojevic2011aerial,risojevic2012orientation} mainly focused on extracting low-level visual attributes and/or modeling mid-level spatial features \cite{kato2014image,lazebnik2006beyond,wang2010locality}. Recently, deep networks, especially CNNs, have achieved a large development in aerial scene recognition~\cite{nogueira2017towards,castelluccio2015land,cheng2018deep}.
Moreover, some methods were proposed to solve the problem of the limited collection of aerial images by employing more efficient networks~\cite{zhang2015scene,zou2015deep,mou2019relation}. Although these methods have achieved great empirical success, they usually learn scene knowledge from the same modality, {\em i.e.}, image. Different from previous works, this paper mainly focuses on exploiting multiple modalities ({\em i.e.} image and sound) to achieve robust aerial scene recognition performance.

\paragraph{Multimodal Learning.}
Information in the real world usually comes as different modalities, with each modality being characterized by very distinct statistical properties, {\em e.g.}, sound and image~\cite{baltruvsaitis2018multimodal}. An expected way to improve relevant task performance is by integrating the information from different modalities. In past decades, amounts of works have developed promising methods on the related topics, such as reducing the audio noise by introducing visual lip information for speech recognition~\cite{hu2016temporal,assael2016lipnet}, improving the performance of facial sentiment recognition by resorting to the voice signal~\cite{zheng2018emotionmeter,zheng2018emotionmeter}.
Recently, more attention is paid to the task of learning to analyze real-world multimodal scenarios~\cite{zhao2018sound,owens2018audio,hu2019deep,hu2020curriculum} and events~\cite{tian2018audio,xiao2020audiovisual}. These works have confirmed the advantages of multimodal learning. In this paper, we proposed to recognize the aerial scene by leveraging the bridge between scene and sound to help better understand aerial scenes.

\paragraph{Cross-task Transfer.} Transferring the learned knowledge from one task to another related task has been approved as an effective way for better data modeling and messages correlating ~\cite{ehrlich2016facial,aytar2016soundnet,imoto2020sound}. 
Aytar et al.~\cite{aytar2016soundnet} proposed a teacher-student framework that transfers the discriminative knowledge of visual recognition to the representation learning task of sound modality via minimizing the differences in the distribution of categories.
Imoto et al.~\cite{imoto2020sound} proposed a method for sound event detection by transferring the knowledge of scenes with soft labels. 
Gan et al.~\cite{gan2019self} transferred the visual object location knowledge for auditory localization learning.
Salem et al.~\cite{salem2018soundscape} proposed to transfer the sound clustering knowledge to the image recognition task by predicting the distribution of sound clusters from an overhead image, similarly work can be found in~\cite{owens2018ambient}.
By contrast, this paper strives to exploit effective sound event knowledge to facilitate the aerial scene understanding task.

\section{Dataset}
\label{section:dataset}

\begin{figure}[t]
	\centering
	\includegraphics[width = 0.9\linewidth]{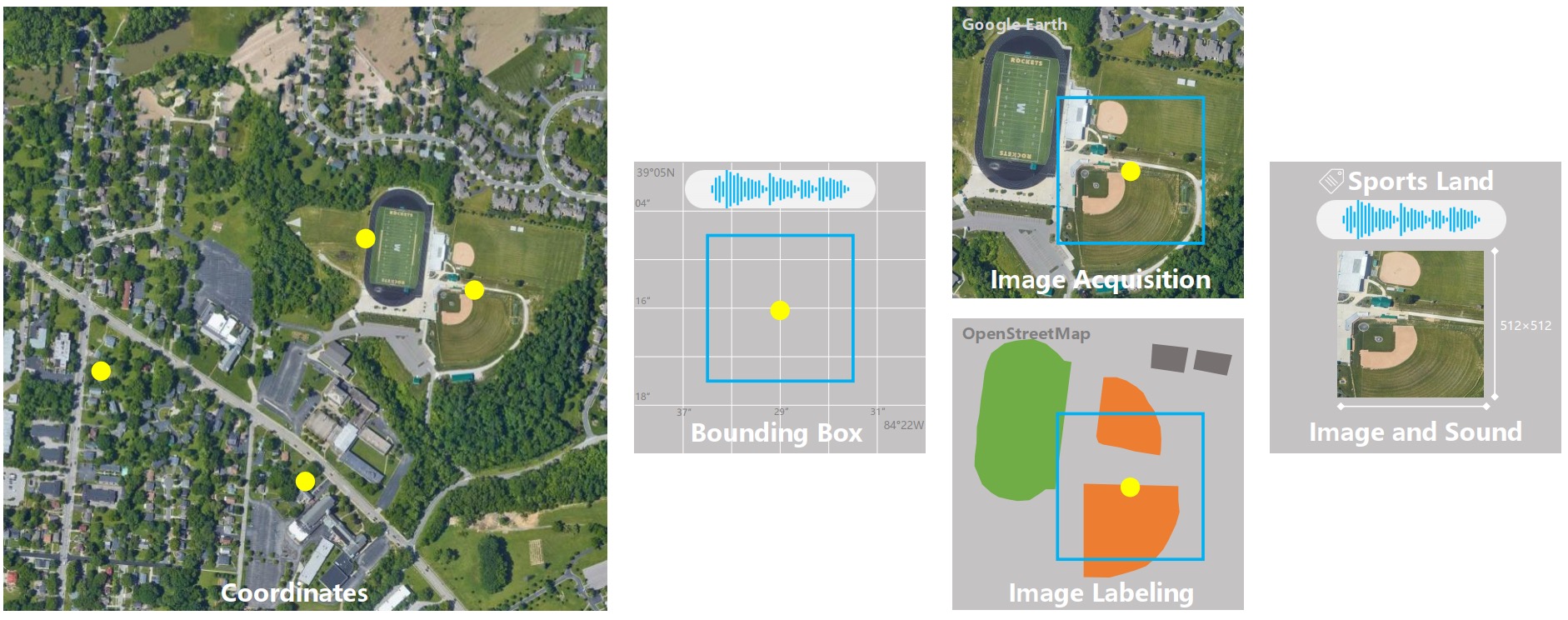}
	\caption{The aerial images acquisition and labeling steps.}
	\label{diagram3}
\end{figure}

To our knowledge, the audiovisual aerial scene recognition task has not been explored before.
Salem {\em et al.}~\cite{salem2018soundscape} established a dataset to explore the correlation between geotagged sound clips and overhead images.
For further facilitating the research in this field, we construct a new dataset, with high-quality images and scene labels, named as ADVANCE\footnote{The dataset webpage: https://akchen.github.io/ADVANCE-DATASET/}, which in summary contains 5075 pairs of aerial images and sounds, classified into 13 classes. 


The audio data are collected from Freesound\footnote{\url{https://freesound.org/browse/geotags/}}, where we remove the audio recordings that are shorter than 2 seconds, and extend those that are between 2 and 10 seconds to longer than 10 seconds by replicating the audio content.
Each audio recording is attached to the geographic coordinates of the sound being recorded.
From the location information, we can download the updated aerial images from Google Earth\footnote{\url{https://earthengine.google.com/}}.
Then we pair the downloaded aerial image with a randomly extracted 10-second sound clip from the entire audio recording content.
Finally, the paired data are labeled according to the annotations from OpenStreetMap\footnote{\url{https://www.openstreetmap.org/}}, also using the attached geographic coordinates from the audio recording.
Those annotations have manually been corrected and verified by participants in case that some of them are not up to date.
The overview of the establishment is shown in Fig.~\ref{diagram3}.

\begin{figure}[t]
	\centering
	\includegraphics[width = 1\linewidth]{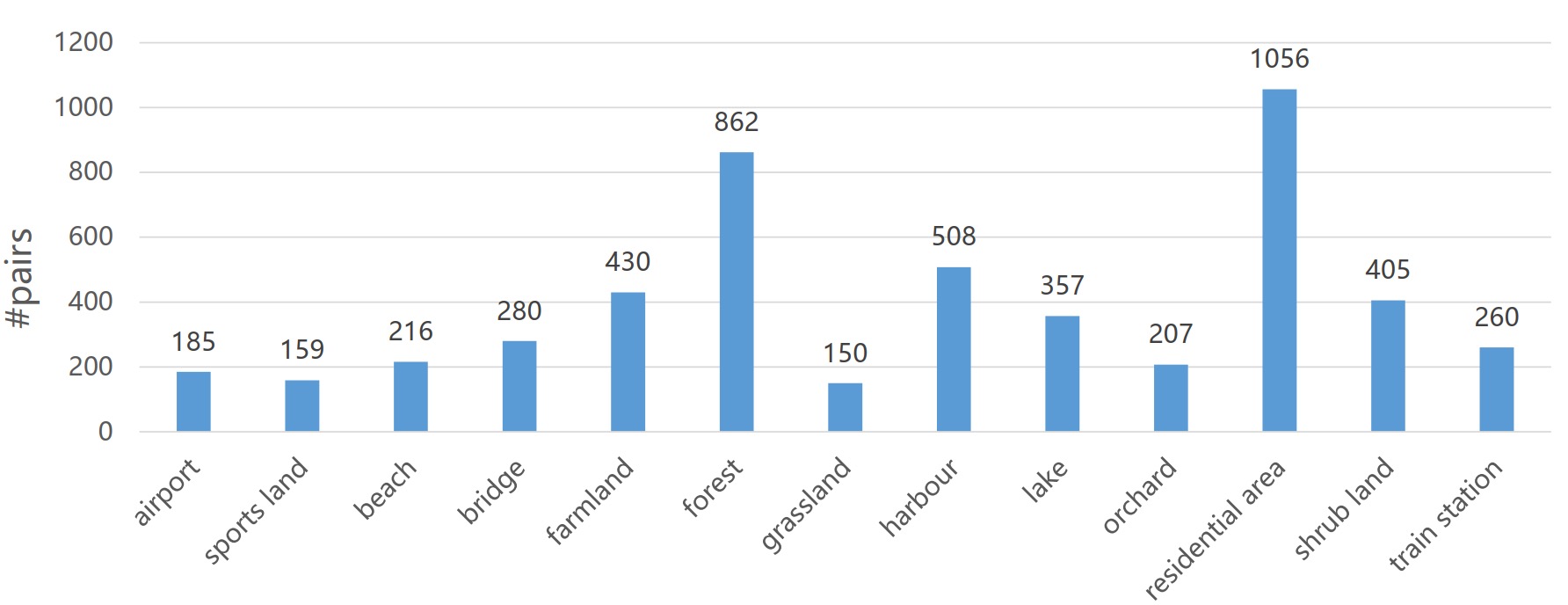}
	\caption{Number of data pairs per class.}
	\label{pairs}
\end{figure}

\begin{figure}[t]
	\centering
	\includegraphics[width = 1\linewidth]{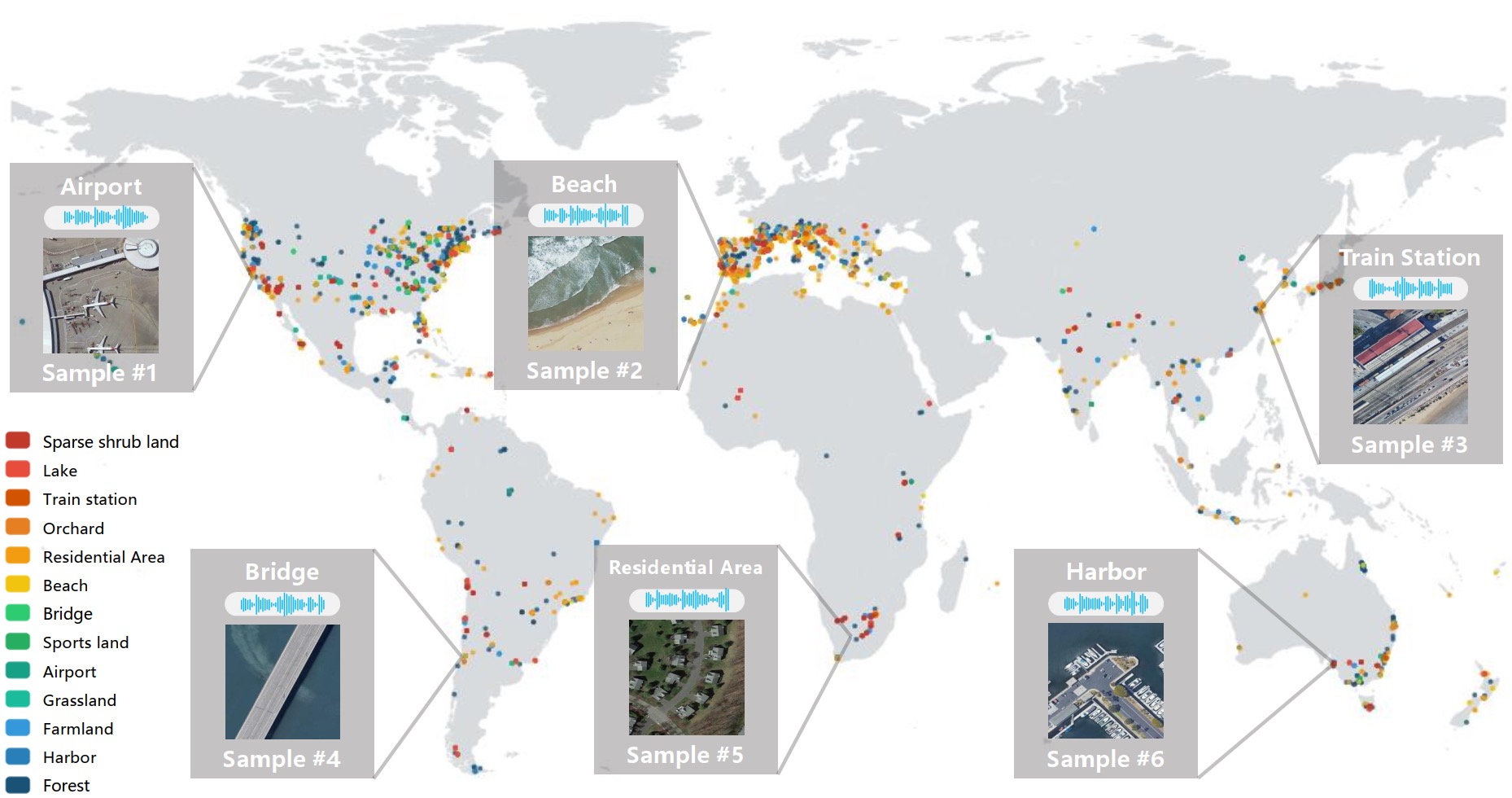}
	\caption{Coordinates distribution and sample pairs of images and sound. Different scenes are represented by different color. Six sample pairs are displayed, which are composed of aerial images, sound and semantic labels. }
	\label{figure_distribution6}
\end{figure}

Due to the inherent uneven distribution of scene classes, the collected data are strongly unbalanced, which makes difficult the training process.
So, two extra steps are designed to alleviate the unbalanced-distribution problem.
Firstly we filter out the scenes whose numbers of paired samples are less than 10, such as desert and the site of wind turbines.
Then for scenes that have less than 100 samples, we apply a small offset to the original geographic coordinates in four directions.
So, correspondingly, four new aerial images are generated from Google Earth and paired with the same audio recording, while for each image, a new 10-second sound clip is randomly extracted from the recording.
Fig.~\ref{pairs} reveals the final number of paired samples per class. 
Moreover, as shown in Fig.~\ref{figure_distribution6}, the samples are distributed over the whole world, increasing the diversity of the aerial images and sounds.

\section{Methodology}
\label{section:method}

In this paper, we focus on the audiovisual aerial scene recognition task, based on two modalities, {\em i.e.}, image and audio. We propose to exploit the audio knowledge to better solve the aerial scene recognition task.
In this section, we detail our proposed approaches for creating the bridge of knowledge transfer from sound event knowledge to the scene recognition in a multi-modality framework.

\begin{table}[htbp]\caption{Main notations.}
\def\arraystretch{1.15}
\begin{center}
\begin{tabular}{p{0.6in} p{4in}}
\toprule
$\avec$,$\vvec$ & audio input, visual input \\
$\xvec$,t & paired image and sound clip, $\xvec=\{\vvec, \avec\}$, and the labeled ground truth $t$ for aerial scene classification\\
$\net_*$ & network, which can be one of the network for extracting visual representation, the network for extracting audio representation, the pretrained (fixed) one for extracting audio representation, {\em i.e.}, $\{\net_v,\net_a,\net_a^{(0)}\}$; also the one that concatenates $\net_v$ and $\net_a$, {\em i.e.} $\net_{v+a}$\\ 
$f_*$ & classifier, which can be one of $\{f_s,f_e\}$, for aerial scene classification or sound event recognition; $f_*$ takes the output of the network as input, and predicts the probability of the corresponding recognition task\\
$\svec$,$\evec$ & probability distribution over aerial scene classes and sound event classes\\
$s_k$,$s_t$& $k$-th scene class' probability, and the $t$-th class being the ground truth\\
$e_k$ & $k$-th sound event class' probability \\
$C(p,q)$ & binary KL divergence: $\log (\frac{p}{q}) + (1-p) \log (\frac{1-p}{1-q})$ \\
\bottomrule
\end{tabular}
\end{center}
\label{tab:notation}
\end{table}

We take the notations from Table \ref{tab:notation}, note that the data $\xvec$ follows the empirical distribution $\mu$ of our built dataset ADVANCE.
For the multimodal learning task with deep networks, we adopt the model architecture that concatenates representations from two deep convolutional networks on images and sound clips.
So our main task, which is a supervised learning problem for aerial scene recognition, can be written as\footnote{For all loss functions, we omit the softmax activation function in $f_s$, the sigmoid activation function in $f_e$, and the expectation of $(\xvec,t)$ over $\mu$ for clarity.}
\begin{equation}
\label{eq:scene-loss}
L_s = -\log \left[f_s(\xvec, \net_{v+a}) \right]_t\enspace,
\end{equation}
which is a cross-entropy loss with $t$-th class being the ground truth.

Furthermore, pre-training on related datasets helps accelerate the training process and improving the performance on the new dataset, especially on a relatively small dataset.
For our task, the paired data samples are limited, and our preliminary experiments show that the two networks $\visionnet$ and $\audionet$ benefit a lot from pre-training on the AID dataset \cite{xia2017aid} for classifying scenes from aerial images, and AudioSet \cite{gemmeke2017audio} for recognizing 527 audio events from sound clips~\cite{wang2018polyphonic}.

\begin{figure*}[t]
	\centering
	\includegraphics[width = \textwidth]{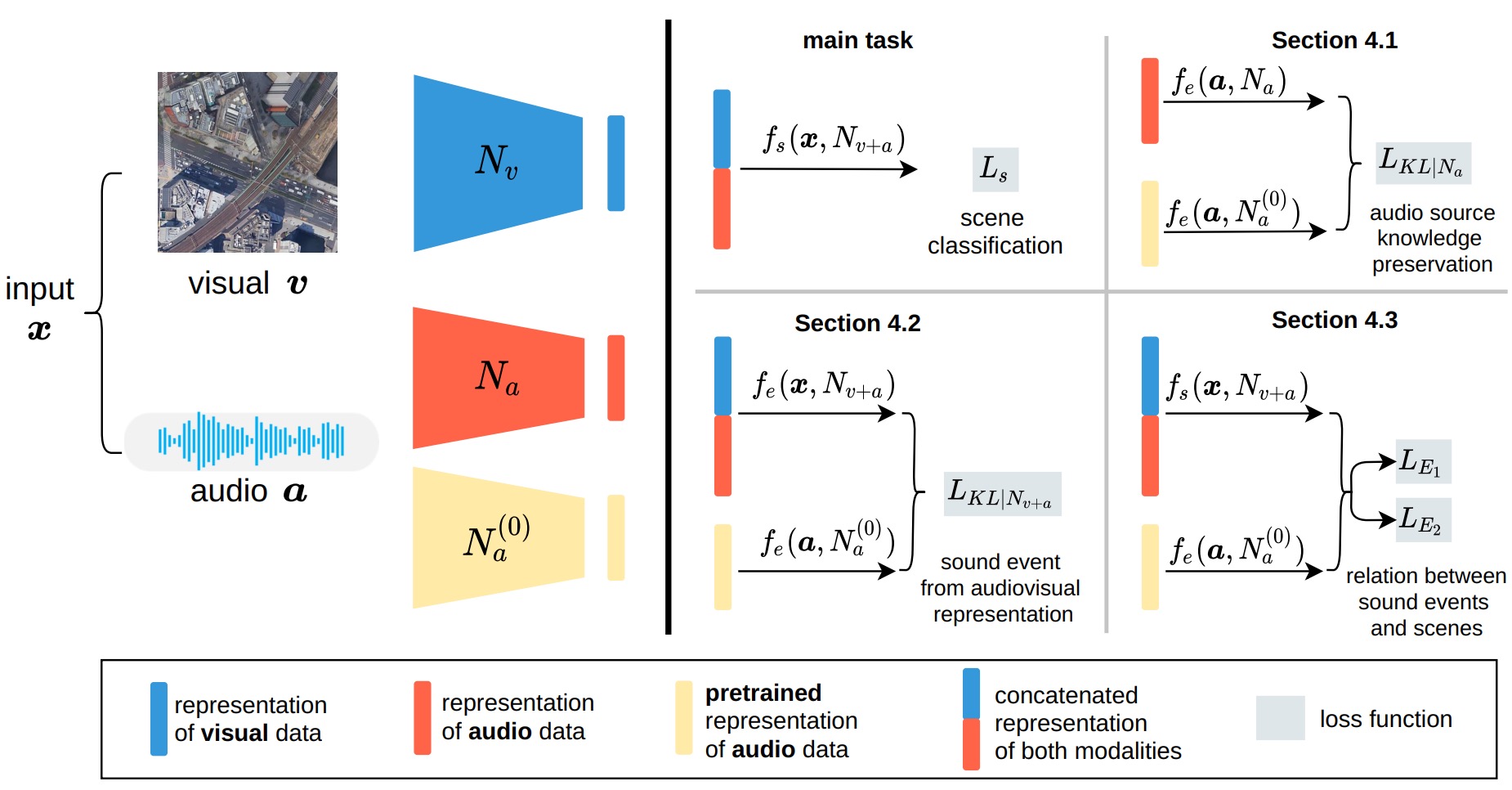}
	\caption{
		Illustration of the main task and three cross-task transfer approaches (best viewed in color).
		We recall the notations:
		$N_v$, with trainable parameters, extracts visual representations, pretrained on the AID dataset; $N_a$, also with trainable parameters, extracts audio representations, pretrained on the AudioSet dataset; $N_a^{(0)}$, is the same as $N_a$ except parameters being fixed; $N_{v+a}$ simply applies both $N_v$ and $N_a$.
		The classifier at the last layer of the network is presented by $f_{task}(input\ data, network)$, where the choice of $task$ is $\{s: \text{scene classification}, e:\text{sound event recognition}\}$, $input\ data$ is one of $\{\vvec, \avec, \xvec\}$, and the set for $network$ is $\{N_v,N_a,N_a^{(0)},N_{v+a}\}$.
		On the left of this figure, our model takes a paired data sample $\xvec$ of an image $\vvec$ and a sound clip $\avec$ as input, and extracts representations from different combinations of modalities and models (shown in different colors);
		On the right, the top-left block introduces our main task of aerial scene recognition, and the rest three blocks present the three cross-transfer approaches.
	}
	\label{fig:pipeline}
\end{figure*}

In the rest of this section, we formulate our proposed model architecture for addressing the multimodal scene recognition task, and present our idea of exploiting the audio knowledge following three directions:
(1) avoid forgetting the audio knowledge during training by preserving the capacity of recognizing sound events;
(2) construct a mutual representation that solves the main task and the sound event recognition task simultaneously, allowing the model to learn the underlying relation between sound events and scenes;
(3) directly learn the relation between sound events and scenes.
Our total objective function $L$ is
\begin{equation}
L = L_s + \alpha L_{\Omega}\enspace,
\end{equation}
where $\alpha$ controls the force of $L_{\Omega}$, and $L_{\Omega}$ is one of the three approaches that are respectively presented in	 Section \ref{subsection:na}, \ref{subsection:nva} and \ref{subsection:bayes}, as illustrated in Fig.~\ref{fig:pipeline}.


\subsection{Preservation of Audio Source Knowledge}
\label{subsection:na}


For our task of aerial scene recognition, audio knowledge is expected to be helpful since the scene information is related to sound events.
While initializing the network by the pre-trained weights is an implicit way of transferring the knowledge to the main task, the audio source knowledge may easily be forgotten during fine-tuning.
Without audio source knowledge, the model can hardly recognize the sound events, leading to a random confusion between sound events and scenes.

For preserving the knowledge, we propose to record the soft responses of target samples from the pre-trained model and retain them during fine-tuning.
This simple but efficient approach \cite{Hinton2015distilling} is named as knowledge distillation, for distilling the knowledge from an ensemble of models to a single model, and has also been used in domain adaptation \cite{tzeng2015simultaneous} and lifelong learning \cite{li2017blearning}.
All of them encourage to preserve the source knowledge by minimizing the KL divergence between the responses from the pre-trained model and the training model.
For avoiding the saturated regions of the softmax, the pre-activations are divided by a large scalar, called temperature \cite{Hinton2015distilling}, to provide smooth responses, with which the knowledge can be easily transferred.

However, for the reason that the audio event task is a multi-label recognition, $f_e(\xvec, \net_*)$ is activated by the sigmoid function. 
The knowledge distillation technique is thus implemented by a sum of binary Kullback-Leibler divergences:
\begin{equation}
\label{eq:event-loss}
L_{KL|\net_a} = \sum_{i} C(\ [f_e(\avec, \net_a^{(0)})]_i\ ||\ [f_e(\avec, \net_a)]_i\ ) \enspace,
\end{equation}
where $[f_e(\avec, \net_*)]_i$ indicates the probability of $i$-th sound event happening in sound clip $\avec$, predicted by $\net_a^{(0)}$ or $\net_a$.
This approach helps to preserve the audio knowledge from the source pretrained network from the AID dataset.



\subsection{Audiovisual Representation for Multi-Task}
\label{subsection:nva}

Different from the idea of preserving the knowledge within the audio modality, we encourage our multimodal model, along with the visual modality, to learn a mutual representation that recognizes scenes and sound events simultaneously.
Specifically, we optimize to solve the sound event recognition task using the concatenated representation, with the knowledge distillation technique:
\begin{equation}
\label{eq:event-loss-concat}
L_{KL|\net_{v+a}} = \sum_{i} C(\ [f_e(\avec, \net_a^{(0)})]_i\ ||\ [f_e(\xvec, \net_{v+a})]_i\ ) \enspace.
\end{equation}

This multi-task technique is very common within one single modality, such as solving depth estimation, surface normal estimation and semantic segmentation from one single image \cite{eigen2015predicting}, or recognizing acoustic scenes and sound events from audio \cite{imoto2020sound}.
We apply this idea to multi-modality, and implement with Equation \eqref{eq:event-loss-concat}, encouraging the multimodal model to learn the underlying relationship between the sound events and the scenes for solving the two tasks simultaneously.

Knowledge distillation with high temperature is equivalent to minimizing the squared Euclidean distance (SQ) between the pre-activations \cite{Hinton2015distilling}.
Instead of minimizing the sum of binary KL divergences, we also propose to directly compare the pre-activations from the networks.
Thereby, we also evaluate $L_{SQ}$ variant for Equation \ref{eq:event-loss} and \ref{eq:event-loss-concat} respectively:
\begin{equation}
\begin{split}
L_{SQ|\net_a} &= \norm[2]{\hat{f}_e(\avec, \audionetinit) - \hat{f}_e(\avec, \audionet)}^2 \enspace,\\
\Check{L}_{SQ|\net_{v+a}} &= \norm[2]{\hat{f}_e(\avec, \audionetinit) - \hat{f}_e(\xvec, \net_{v+a})}^2 \enspace,
\end{split}
\end{equation}
where $\hat{f}_e$ is the pre-activations, recalling that $f_e$ is activated by sigmoid.

\subsection{Sound Events in Different Scenes}
\label{subsection:bayes}

The two previously proposed approaches are based on the multi-task learning framework, either using different or the same representations, in order to preserve the audio source knowledge or implicitly learn an underlying relation between aerial scenes and sound events.
Here, we propose an explicit way for directly modeling the relation between scenes and sound events, and creating the bridge of transferring the knowledge between two modalities.

We employ the paired image-audio data samples from our built dataset as introduced in Section \ref{section:dataset}, analyze the happening sound events in each scene, and obtain the posteriors given one scene.
Then instead of predicting the probability of sound events by the network, we estimate this probability distribution $p(\evec)$ with the help of posteriors $p(\evec | s_k)$ and the predicted probability of scenes $p(\svec)$:
\begin{equation}
\label{eq:posteriors}
p(\evec) = \sum_k \ p(s_k) \ p(\evec | s_k) = \sum_k \ [f_s(\xvec, \net_{v+a})]_k \ p(\evec | s_k)\enspace,
\end{equation}
where $p(s_k)=[f_s(\xvec, \net_{v+a})]_k$ is the predicted probability of the $k$-th scene, and the posteriors $p(\evec | s_k)$ is obtained by averaging $f_e(\avec, \net_a^{(0)})$ over all samples that belong to the scene $s_k$.
This estimation $p(\evec)$ is in fact the compound distribution that marginalizes out the probability of scenes, while we search for the optimal scene probability distribution $p(\svec)$ (ideally one-hot) through aligning $p(\evec)$ with soft responses:
\begin{equation}
\label{eq:bayes-loss-1}
L_{E_1} = \sum_{i} C(\ [f_e(\avec, \net_a^{(0)})]_i\ ||\ p(e_i)\ ) \enspace.
\end{equation}

Besides estimating the probability of each sound event happening in a specific scene, we also investigate possible concomitant sound events.
Some sound events may largely overlap under a given scene, and this coincidence can be used as a characteristic for recognizing scenes.
We propose to extract this characteristic from $f_e(\avec, \net_a^{(0)})$ of all audio samples that belong to this specific scene.

We note $P(\evec|s_k) \in \Rset^{n_k \times c}$ as the sound event probabilities of $n_k$ samples in the scene $s_k$, where each row is each sample's probability of sound events in the scene $s_k$.
Then with the Gram matrix $P(\evec|s_k)^T P(\evec|s_k)$, we extract the largest eigenvalue and the corresponding eigenvector $\dvec_k$ as the characteristic of $P(\evec|s_k)$.
This eigenvector $\dvec_k$ indicates the correlated sound events and quantifies their relevance in the scene $s_k$ by the direction of this vector.
We thus propose to align the direction of $\dvec_t$, the event relevance of the ground truth scene $s_t$, with the estimated $p(\evec)$ from Equation \ref{eq:posteriors}:
\begin{equation}
\label{eq:bayes-loss-2}
L_{E_2} =  \text{cosine}( \dvec_t, p(\evec) ) \enspace.
\end{equation}

Equation \eqref{eq:bayes-loss-1} and \eqref{eq:bayes-loss-2} have provided a way of explicitly building the connection between scenes and sound events.
In the experiments, we use them together:
\begin{equation}
\label{eq:bayes-loss}
L_{E} = L_{E_1} + \beta L_{E_2}\enspace,
\end{equation}
where $\beta$ is a hyper-parameter controlling the importance of $L_{E_2}$.

\section{Experiments}
\label{section:exp}

\subsection{Implementation Details}
Our built ADVANCE dataset is employed for evaluation, where 70\% image-sound pairs are for training, 10\% for validation, and 20\% for testing. Note that, these three sub-sets do not share audiovisual pairs that are collected from the same coordinate. Before feeding the recognition model, we sub-sample the sound clips at 16 kHz. Then, following \cite{wang2018polyphonic}, the short-term Fourier transform is computed using a window size of 1024 and a hop length of 400. The generated spectrogram is then projected into the log-mel scale to obtain an audio matrix in $\Rset^{T\times F}$, where the time $T=400$ and the frequency $F=64$. Finally, we normalize each feature dimension to have zero mean and unit variance. The image data are all resized into 256$\times$256, and horizontal flipping, color, and brightness jittering are used as data augmentation means. 

In the network setting part, the visual pathway employs the AID pre-trained ResNet-101 for modeling the scene content~\cite{xia2017aid} and the audio pathway adopts the AudioSet pre-trained ResNet-50 for modeling the sound content~\cite{wang2018polyphonic}. The whole network is optimized via an Adam optimizer with a weight decay rate 1e-4 and a relatively small learning rate 1e-5, as both backbones have been pre-trained from external knowledge. By using grid search strategy, the hyper-parameters of $\alpha$ and $\beta$ are set as 0.1 and 0.001, respectively. We adopt the weighted-averaging precision, recall and F-score metrics for evaluation, which are more convincing when faced with uneven distribution of scene classes.

\subsection{Aerial Scene Recognition}

\begin{figure}[!ht]
\makebox[\textwidth]{
	\scalebox{0.6}{\input{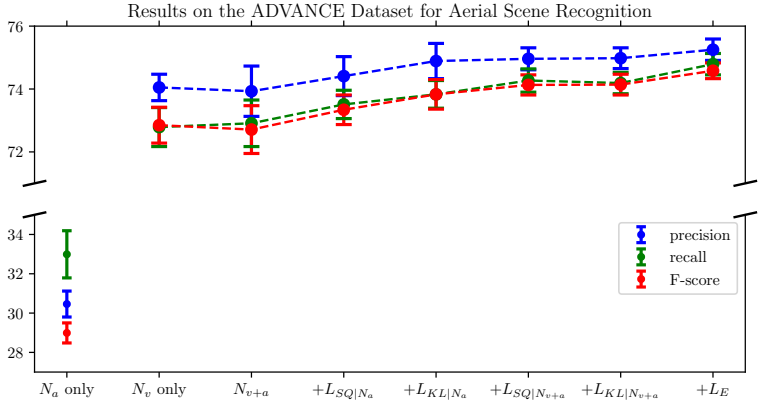}}
}
\caption{Aerial scene recognition results on the ADVANCE dataset from 5 different runs, where the first approaches perform only the main loss function $L_s$, and the approaches with the symbol $+$ mean they are respectively combined with $L_{s}$.}
\label{fig:scene_class}
\end{figure}

Fig.~\ref{fig:scene_class} shows the recognition results of different learning approaches under the unimodal and multimodal scenario, from which we have four points should pay attention to. Firstly, according to the unimodal results, the sound data can provide a certain reference for different scene categories, although it is significantly worse than image-based results. Such phenomenon reminds us that we can take advantage of the audio information to improve recognition results further. Secondly, we recognize that simply using the information from both modalities does not bring benefits but slightly lowers the results (72.85 vs. 72.71 in F-score). This could be because the pre-trained knowledge for audio modality may be forgotten or the audio messages are not fully exploited just with the rough scene labels. Thirdly, when the sound event knowledge is transferred for the scene modeling, we have considerable improvements for all of the proposed approaches. The results of $L_{SQ|N_a}$ and $L_{KL|N_a}$ show that preserving audio event knowledge is an effective means for better exploiting audio messages for scene recognition, and the performance of $L_{SQ|N_{v+a}}$ and $L_{KL|N_{v+a}}$ demonstrates that transferring the unimodal knowledge of sound events to the multimodal network can help to learn better mutual representation of scene content across modalities.
Fourthly, among all the compared approaches, our proposed $L_{E}$ approach shows the best results, as it better imposes the sound event knowledge by imposing the underlying relation between scenes and sound events. 

\begin{figure}[t]
	\centering
	\includegraphics[width=11cm]{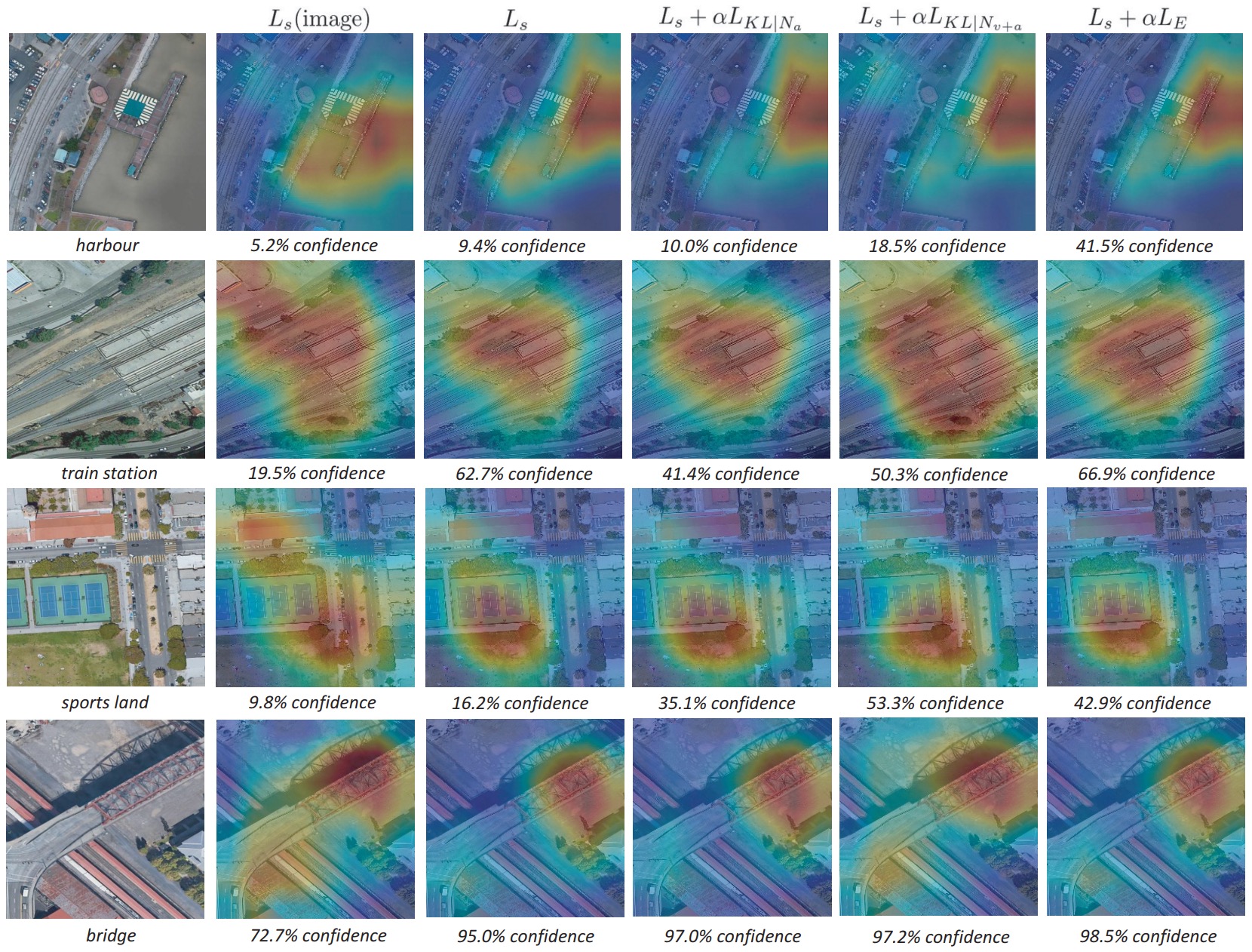}
	\caption{The class activation map generated by different approaches for different categories, as well as the corresponding predict probabilities of ground-truth category. $L_s$(image) means the learning objective of $L_s$ is just performed with image data.}
	\label{fig:cam}
\end{figure}

We use the CAM technique \cite{zhou2016learning} to highlight the parts of the input image that make significant contributions to identifying the specific scene category. Fig.~\ref{fig:cam} shows the comparison of the visualization results and  the predicted probabilities of the ground-truth label among different approaches. By resorting to the sound event knowledge, as well as its association with scene information, our proposed model can better localize the salient area of the correct aerial scene and provide a higher predicted probability for the ground-truth category, e.g, the \emph{harbour} and \emph{bridge} class.
s

Apart from the multimodal setting, we have also conducted more experiments under the unimodal settings, shown in Table~\ref{table:unimodal-result}, for presenting the contributions from pre-trained models, and verifying the benefits from the sound event knowledge on the aerial scene recognition.
For these unimodal experiments, we keep one modal data input and set the other to zeros.
When only the audio data are considered, the sound event knowledge is transferred within the audio modality and thus $L_{SQ|N_{a}}$ is equivalent to $L_{SQ|N_{v+a}}$, similarly for the visual modality case.
Comparing the results of randomly initializing the weights {\em i.e.} $L_s^\dagger$ and initializing with the pre-trained weights {\em i.e.} $L_s$, we find that initializing the network from the pre-trained model can significantly prompt the performance, which confirms that pre-training from a large-scale dataset benefits the learning task on the small datasets.
Another remark from this table is that the results of the three proposed approaches show that both unimodal networks can take advantage of the sound event knowledge to achieve better scene recognition performance.
It further validates the generalization of the proposed approaches, either in the multimodal or the unimodal input case.
Compared with the multi-task framework of $L_{SQ|N_{v+a}}$ and $L_{KL|N_{v+a}}$, the $L_{E}$ approach can better utilize the correlation between sound event and scene category via the statistical posteriors.

\begin{table}
	\begin{center}
		\caption{Unimodal aerial scene recognition results on the ADVANCE dataset from 5 different runs, where $\dagger$ means random initialization and the approaches with the symbol $+$ mean they are weightedly combined with $L_{s}$.}
		\label{table:unimodal-result}
		\begin{tabular}{c|c|c|c|c|c|c}
			\hline\noalign{\smallskip}
			Modality & Approaches & ${L_s}^\dagger$ & $L_s$ & $+L_{SQ|N_{v+a}}$ & $+L_{KL|N_{v+a}}$ & $+L_{E}$ \\
			\noalign{\smallskip}
			\hline
			\noalign{\smallskip}
			\multirow{3}*{Sound}& Precision & 21.15$\pm$0.68& 30.46$\pm$0.66 & 31.64$\pm$0.65&30.00$\pm$0.86&31.14$\pm$0.30 \\
			& Recall      &24.54$\pm$0.67 & 32.99$\pm$1.20 & 34.68$\pm$0.49&34.29$\pm$0.35&33.80$\pm$1.03\\
			& F-score    & 21.32$\pm$0.42 & 28.99$\pm$0.51 & 29.31$\pm$0.71&28.51$\pm$0.99&29.66$\pm$0.13\\
			\noalign{\smallskip}
			\hline
			\noalign{\smallskip}
			\multirow{3}*{Image}& Precision & 64.45$\pm$0.97 &74.05$\pm$0.42 &74.86$\pm$0.94 &74.36$\pm$0.85&73.97$\pm$0.39 \\
			& Recall      & 64.59$\pm$1.12 &72.79$\pm$0.62 &74.11$\pm$0.89 &73.40$\pm$0.84&73.47$\pm$0.52\\
			& F-score    & 64.04$\pm$1.07 &72.85$\pm$0.57 &73.98$\pm$0.92 &73.52$\pm$0.85&73.44$\pm$0.45\\
			\noalign{\smallskip}
			\hline
		\end{tabular}
	\end{center}
\end{table}

\subsection{Ablation Study}
In this subsection, we directly validate the effectiveness of the scene-to-event transfer term $L_{E_1}$ and the event relevance term $L_{E_2}$, without the supervision from the scene recognition objective of $L_s$. Table~\ref{table:scene-event} shows the comparison results. By resorting to the scene-to-event transfer term, performing sound event recognition can reward the model the ability to distinguish different scenes. When further equipped with the event relevance of the scenes, the model can have higher performance.
This demonstrates that cross-task transfer can indeed provide reasonable knowledge if the inherent correlation between these tasks are well exploited and utilized. By contrast, as the multi-task learning approaches do not well take advantage of this knowledge, the scene recognition performance remains at the chance level.

\begin{figure}[t]
	\centering 
	\subfigure[$L_s + \alpha L_{SQ|N_a}$]{ 
		\includegraphics[width=1.6in]{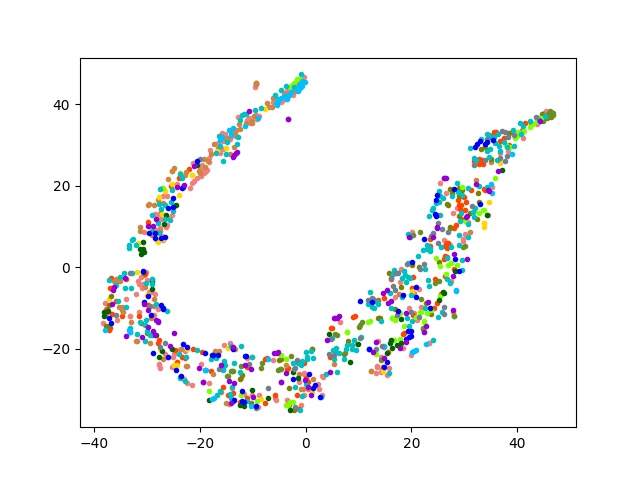} 
	} \hspace{-8mm}
	\subfigure[$L_s + \alpha L_{SQ|N_{v+a}}$]{ 
		\includegraphics[width=1.6in]{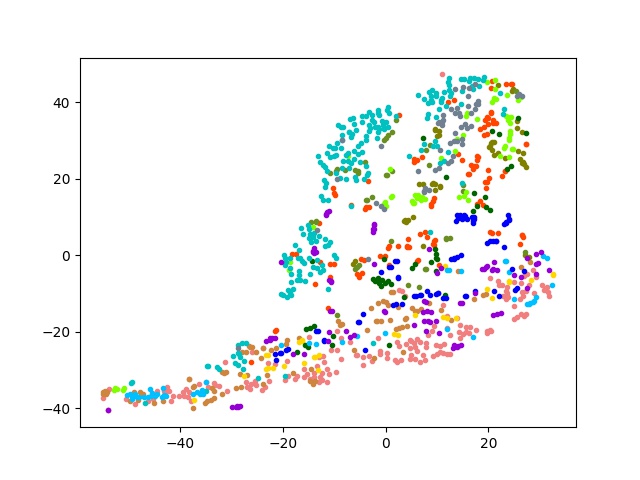} 
	} \hspace{-8mm}
	\subfigure[$L_s + \alpha L_{E}$]{ 
		\includegraphics[width=1.6in]{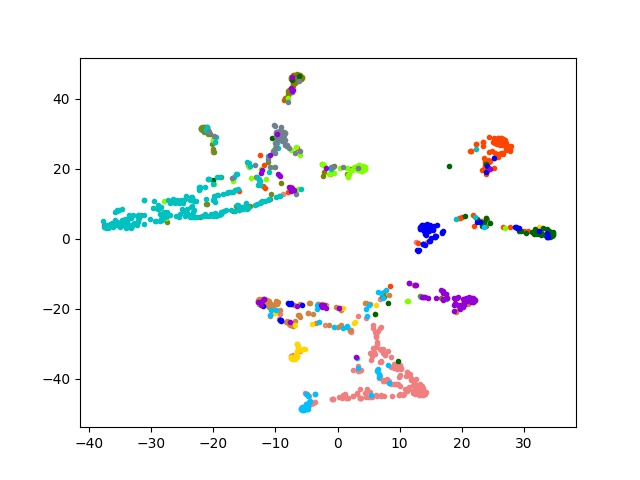} 
	} \hspace{-8mm}
	\caption{The aerial scene data embeddings indicated by the corresponding sound event distribution, where the points in different color mean in different scene categories.}
	\label{fig:tsne}
\end{figure}

\begin{table}[h]
	\begin{center}
		\caption{Aerial scene recognition results on the ADVANCE dataset, where only the sound event knowledge is considered in the training stage.}
		\label{table:scene-event}
		\begin{tabular}{c|c|c|c|c}
			\hline\noalign{\smallskip}
			Approaches & $L_{E_1}$ & $L_{E_1}+\beta L_{E_2}$ & $L_{KL|N_{v+a}}$ & $L_{SQ|N_{v+a}}$\\
			\noalign{\smallskip}
			\hline
			\noalign{\smallskip}
			Precision & 43.37$\pm$0.59& 54.23$\pm$1.14	& 3.08$\pm$0.14	& 2.95$\pm$0.07\\
			Recall      &49.26$\pm$0.36 & 52.57$\pm$0.72 & 9.69$\pm$0.43 & 9.28$\pm$0.17 \\
			F-score    & 42.50$\pm$0.42 & 48.65$\pm$0.85 & 4.46$\pm$0.20 & 4.24$\pm$0.07\\
			\noalign{\smallskip}
			\hline
		\end{tabular}
	\end{center}
\end{table}

To better illustrate the correlation between aerial scenes and sound events, we further visualize the embedding results. Specifically, we use the well-trained cross-task transfer model to predict the sound event distribution on the testing set. Ideally, the sound event distribution can separate the scenes from each other, since each scene takes a different sound event distribution. Hence, we use t-SNE~\cite{maaten2008visualizing} to visualize the high-dimensional sound event distributions of different scenes. Fig.~\ref{fig:tsne} shows the visualization results, where the points in different color mean different scene categories. As $L_{SQ|N_a}$ is performed within the audio modality, the sound event knowledge cannot well transfer to the entire model, leading to the mixed scene distribution. By contrast, as $L_{SQ|N_{v+a}}$ transfers the sound event knowledge into the multimodal network, the predicted sound event distribution can separate different scenes to some extent. By introducing the correlation between scenes and events, {\em i.e.}, $L_{E}$, different scenes can be further disentangled, which confirms the feasibility and merits of cross task transfer.

\section{Conclusions}
In this paper, we explore a novel multimodal aerial scene recognition task that considers both visual and audio data.
We have constructed a dataset consists of labeled paired audiovisual worldwide samples for facilitating the research on this topic.
We propose to transfer the sound event knowledge to the scene recognition task for the reasons that the sound events are related to the scenes and that this underlying relation is not well exploited. Amounts of experimental results show the effectiveness of three proposed transfer approaches, confirming the benefit of exploiting the audio knowledge for the aerial scene recognition. 

\section{Acknowledgement}
This work was supported in part by the National Natural Science Foundation of China under Grant 61822601 and 61773050; the Beijing Natural Science Foundation under Grant Z180006.

\clearpage
%
%
\bibliographystyle{splncs04}
\bibliography{egbib}
\end{document}